# Dual-Individual Genetic Algorithm: A Dual-Individual Approach for Efficient Training of Multi-Layer Neural Networks

**Tran Thuy Nga Truong[1] and Jooyong Kim***

* Department of Materials Science and Engineering;

***** Correspondence: jykim@ssu.ac.kr; Tel.: +82-10-8720-0631

*[1] Department of Smart Wearables Engineering*

*Soongsil University, Seoul 156-743, Korea*

**Abstract:** This paper introduces an enhanced Genetic Algorithm technique, which optimizes neural networks for binary image classification tasks, such as cat vs. non-cat classification. The proposed method employs only two individuals for crossover, represented by two parameter sets: Leader and Follower. The Leader focuses on exploitation, representing the primary optimal solution, while the Follower promotes exploration by preserving diversity and avoiding premature convergence. Leader and Follower are modeled as two phases or roles. The key contributions of this work are threefold: (1) a self-adaptive layer dimension mechanism that eliminates the need for manual tuning of layer architectures; (2) generates two parameter sets, leader and follower parameter sets, with 10 layer architecture configurations (5 for each set), ranked by Pareto dominance and cost post-optimization; and (3) achieved better results compared to gradient-based methods. Experimental results show that the proposed method achieves 99.04% training accuracy and 80% testing accuracy (cost = 0.06) on a three-layer network with architecture [12288, 17, 4, 1], higher performance a gradient-based approach that achieves 98% training accuracy and 80% testing accuracy (cost = 0.092) on a four-layer network with architecture [12288, 20, 7, 5, 1].

*Corresponding author: jykim@ssu.ac.kr

[1] Author: thuynga290391@gmail.com

## 1. INTRODUCTION

Reinforcement Learning (RL) is the strategy of learning where an agent learns optimal behaviors by interacting with an environment through trial and error. The agent performs actions, receives rewards or penalties as feedback, and aims to maximize the cumulative reward over time [1]. RL has made exciting progress in domains like game playing (e.g., AlphaGo), robotics, and autonomous systems. However, it still faces challenges, such as sparse rewards[2,3], high-dimensional action spaces [4], and training instability [5]. Genetic Algorithms (GA), inspired by the principles of natural evolution, such as selection, mutation, and reproduction, offer versatile support for RL across multiple stages [6]. GAs excel at optimizing hyperparameters, such as learning rates and network architectures, or exploration in challenging environments, particularly those with sparse rewards, by maintaining population diversity and preventing agents from getting stuck in suboptimal strategies [7,8]. GAs can also generate a diverse set of initial policies, providing RL agents with a solid starting point that can be further refined through training. GAs operate by maintaining a population of candidate solutions, which evolve over generations through mechanisms analogous to biological evolution. Their ability to explore complex search spaces and handle diverse problem types makes them highly versatile. However, their performance is heavily influenced by factors such as population size, parameter settings, and computational resources [9]. There are two critical issues in the evolution process of the genetic search: population diversity (to avoid premature convergence) and selective pressure (to prioritize high-quality solutions), both of which are critically influenced by population size [10]. A small population risks rapid convergence due to low diversity, while a large population may waste computational resources despite enhancing exploration. GA methods adapt to these trade-offs through varying population scales: Standard GA (100–10,000+ individuals) maximizes global exploration but demands high computational power; Parallel/Island GA (subpopulations of > 64) balances diversity and efficiency via distributed evolution [11,12]; Microbial GA (>10 individuals) focuses on localized adaptation at the risk of early convergence [13]; and (1+1)-Evolution Strategy (1 parent + 1 offspring) enables rapid unimodal optimization at the cost of diversity[14–16]. In the (1+1)-ES, since there's only one parent and one offspring, there's no recombination, only mutation. Only top-tier candidates are selected, meaning the best between parent and offspring is chosen. It's called "plus" because the parent competes with the offspring. Each generation has one parent and one offspring. The parent creates an offspring through mutation, and then they compete. The better one survives to the next generation.

In this paper, we introduce an enhanced GA that uses two individuals for matching, recombination, and mutation, differing from the single-parent (1+1)-ES [17]. In the proposed method, individuals are hierarchically labeled as Leader (possessing the optimal cost value) and Follower (suboptimal), with roles switching if the Follower finds a better solution in subsequent iterations. The concept of Leader-Follower involves a dynamic interchange of roles or states, where the Leader sometimes becomes the Follower and vice versa, representing a state of balance and mimicking natural balancing mechanisms. This dynamic structure maintains constant competition: the Leader drives the exploitation of high-quality solutions, while the Follower promotes exploration by introducing diversity. Unlike standard GAs (which treat parents as interchangeable) or (1+1)-ES (limited by single-parent evolution), the proposed method balances exploitation and exploration through role-based collaboration and recombination, preventing premature convergence. The leader and follower do not negate one another. Instead, they coexist in a symbiotic yet competitive relationship, supporting each other while vying for dominance; neither entity fully dominates the other. This is unlike elimination-based strategies, where individuals eliminate each other to retain only the optimal candidate. As a result, the follower persistently pursues and replaces the leader. During initial iterations, the frequency of role-swapping between leader and follower is high, gradually diminishing as the system converges toward an optimal equilibrium. The proposed model shows a slight improvement in performance compared to gradient descent. For a neural network architecture with two layers (12288, 7, 1), the proposed method achieves 100% accuracy on the training set and 74% on the test set, whereas gradient descent achieves 100% accuracy on the training set and 72% on the test set. The improvement, though modest, underscores the effectiveness of the proposed method integrated into the neural network.

## 2. EXPERIMENTAL

### 2.1. Datasets

The dataset used in this research is from Week 4 of course 1 of the Deep Learning Specialization, created by Andrew Ng, co-founder of Coursera and founder of DeepLearning. AI [18]. This dataset includes 209 training images and 50 test images of cats and non-cat objects, each manually labeled and preprocessed to 64x64 RGB pixels, simplified for educational use. Each image is flattened into a 12,288-dimensional vector (64 height x 64 width x 3 RGB channels) and normalized to the range [0, 1] by scaling the pixel values to 255. Labels are binary (1 = cat, 0 = non-cat) and stored as 1D arrays to align with neural network input requirements. Despite its simplicity, the dataset reflects real-world constraints, such as limited training samples and noisy labels, thereby balancing computational feasibility with real-world relevance. The predefined Deep Neural Network (DNN) structures for two-layer ([12288, 7, 1]) and four-layer networks ([12288, 20, 7, 5, 1]), making it easy to compare with the proposed model using two-layers (12288, 7, 1) and three layers (12288, 17, 4, 1).

### 2.2. L-layer deep neural network

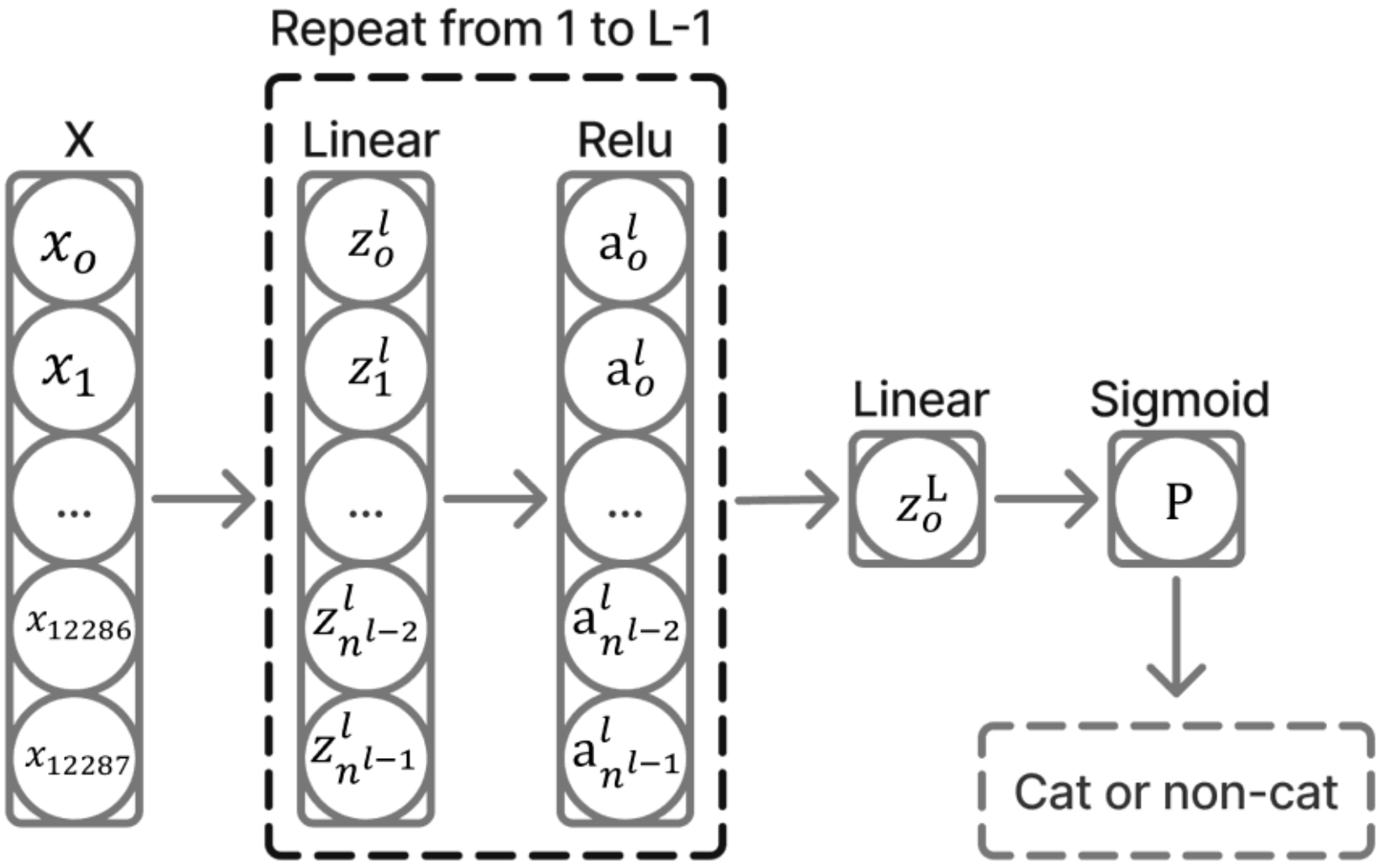

**Figure 1. L-layer Model Neural Network**

In Figure 1, the model structure used in the paper is [LINEAR → RELU] repeated L-1 times, followed by LINEAR → SIGMOID. The input is a 64x64x3 RGB image flattened into a 12288-dimensional vector, which is then multiplied by a weight matrix and added to a bias term, creating a linear unit. ReLU activation is applied, and this repeats for each hidden layer. Finally, a sigmoid layer outputs a probability P for binary classification, cat or not (≥0.5 for "cat" classification). The linear forward module (vectorized over all examples) computes the following equations for each layer:

$$Z^{[l]} = W^{[l]}.A^{[l-1]} + b^{[l]} \quad (1)$$

Where: $A^{[l-1]}$ is the activation matrix from the previous layer (or input matrix X for l=1 or $A^{[0]} = X$ ), with dimensions ($n^{[l-1]}$, m), where m is the number of examples. $W^{[l]}$ is the weight matrix of the current layer, with dimensions ($n^{[l]}$ , $n^{[l-1]}$). $b^{[l]}$ is the bias vector of the current layer, with dimensions ($n^{[l]}$, 1). $Z^{[l]}$ is the linear output matrix of the current layer, with dimensions ($n^{[l]}$ ,m). For example, in a 2-layer network with ReLU and sigmoid activations. The first layer (ReLU) is $Z^{[1]} = W^{[1]}.X + b^{[1]}$ and $A^{[1]} = ReLU(Z^{[1]})$. The second layer (Sigmoid) is $Z^{[2]} = W^{[2]}.A^{[1]} + b^{[2]}$ and $A^{[2]} = \sigma(Z^{[2]})$.

For binary classification, the cross-entropy cost function is typically given by the formula below:

$$J = -\frac{1}{m}\sum_{i=1}^{m}\left[y^{(i)}\log\left(a^{[L](i)}\right) + (1-y^{(i)})\log\left(1-a^{[L](i)}\right)\right] \qquad (2)$$

Where m is the number of training examples, $y^{(i)}$ is the true label (0 or 1), and $a^{[L](i)}$ is the predicted probability from the sigmoid output after propagating through the final output layer L for the i-th example.

## 2.3.Outline the proposed model architecture

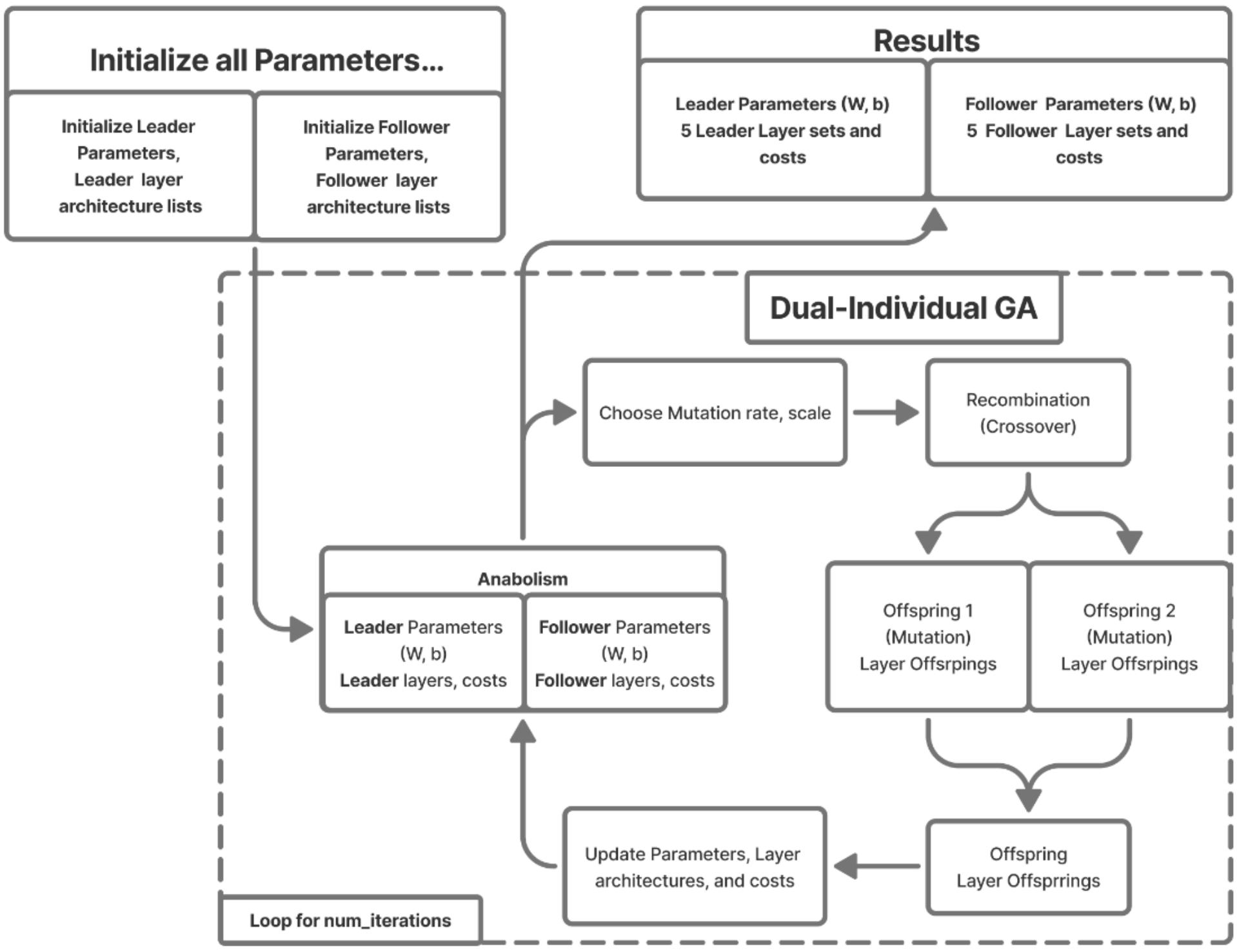

**Figure 2. The framework of the proposed model**

**ALGORITHM 1: DUAL-INDIVIDUAL GA**

***Input:*** *Size , X_train, Y_train, max_layer_dims, stop_cost*

***Output:*** *Best solution, best cost, leader_params, follower_params, leader_layers, follower_layers*

***Initialization of parameters:***

*leader_layers, leader_params =* ***initialize_agent_network****(size, X_train, Y_train, max_layer_dims)*

*follower_layers, follower_params =* ***initialize_agent_network****(size, X_train, Y_train, max_layer_dims)*

***FOR*** *iteration = 0* ***to*** *20,000:*

```
    IF best_cost < stop_cost:
        Return Output
    mutation_rate = linear_value(iteration, max_iter, min_val = 0.1, max_val = 0.9)
    mutation_scale = 0.0008
    off_layers = create_new_solutions(leader_layers, follower_layers, CR, PAR, max_layer_dims)
    off_params_1, off_params_2 = crossover_rows_vectorized(leader_params, follower_params)
    off_params_1, off_params_2 = mutate(off_params_1, off_params_2, mutation_rate, mutation_scale)
    off_layers_1, off_layers_2 = update_and_sort(off_layers, off_params_1, X_train, Y_train)
    off_params, off_layers = merge_two_offs(off_params_1, off_params_2, off_layers_1, off_layers_2)
    off_params, off_layers = update_and_sort(off_layers, off_params, X_train, Y_train)
    leader_params, follower_params, leader_layers, follower_layers = anabolism_params(
            leader_params, follower_params, off_params, leader_layers, follower_layers, off_layers)
    leader_params, follower_params, leader_layers, follower_layers = update_lead_foll(leader_params,
                                follower_params, leader_layers, follower_layers, X_train, Y_train)
    Best_cost = min([solution[-1] for solution in leader_layers])
Return Output
End
```

Figure 2 illustrates the implementation framework of the Dual-Individual Genetic Algorithm, while Algorithm 1 outlines its core optimization workflow. Max_layer_dims represents the maximum possible values that the create_new_solutions function can generate for a new list of layer dimensions. The function generates neural architectures with fixed input and output dimensions (e.g., input = 12288, output = 1) while optimizing hidden layer sizes within predefined bounds. For instance, using max_layer_dims = [12288, 20, 10, 1], the hidden layers (indices 1-2) ≤ 20 and 10 are dynamically adjusted during evolution, but the first and last layers remain unchanged. To ensure consistency, both leader_params and follower_params are initialized to match the dimensions specified in max_layer_dims. The Algorithm Workflow is as follows:

**Initialization:** Leader & Follower Parameters (leader_params, follower_params) and their corresponding layer dimensions agent_layers (leader_layers, follower_layers) are initialized via initialize_agent_network. Layer dimensions are constrained by max_layer_dims, with input/output layers fixed to match the dataset (X_train, Y_train). In this context, size refers to the number of candidate solutions stored in agent_layers. Each solution represents a neural network architecture (layer dimensions) along with its associated cost value. For example, with size = 2, two solutions are stored in the agent_layers formatted as agent_layers = [[12288, 3, 1, 0.43086], [12288, 4, 1, 0.63061]]. Each sub-list defines a 2-layer neural network; for example, [12288, 3, 1, 0.43086] specifies: input = 12288, hidden = 3, output = 1, and cost value = 0.43086. The purpose of size is to control the diversity of architectures explored. A larger size allows for broader exploration, but it also increases computational cost. In this study, a sample size of 5 is chosen. Both agents' architectures (leader_layers, follower_layers) undergo forward propagation to compute their costs. Solutions are sorted by Pareto dominance ranking and cost values.

**Iterative Optimization Loop (max_iter = 20,000 iterations):** The maximum number of iterations is set to 20,000 as an upper bound; however, the algorithm may terminate earlier if the cost falls below the predefined threshold in stop_cost, allowing for comparison of iteration counts across different layer parameter configurations. New Solutions generate offspring layer dimensions (off_layers) via create_new_solutions(). Crossover creates off_params_1 and off_params_2 by crossing over agent_params (leader_params and follower_params) via crossover_rows_vectorized(). Apply the mutation scale and rate to both offspring parameter sets via mutation(). Compute costs for off_layers_1 and off_layers_2 (the same architecture layer with off_layers, but different costs due to different parameter sets) and merge offspring into a unified off_params and off_layers. anabolism_params updates agent_params and their layer dimensions agent_layers using the offspring (off_params and off_layers). This mimics biological "anabolism," where new solutions are synthesized and integrated into agent_params. Updated and compared costs via update_lead_foll() between agent_layers, swap agent_layers if follower_layers has a better cost.

## 2.4.Initialize all parameters

In the proposed method, all initial weights and biases are set to zero based on max_layer_dims. The agent layers have unique architectures by varying hidden layer units, but some might overlap. The key point is that since all agents start with the same parameters, their initial predictions are the same, leading to identical cost values (like 0.6931 for binary classification with sigmoid and log loss). In Table 1, there are 2 agents, each initialized with 5 different neural network architectures. After the first training update, the divergence between the Leader and Follower agents occurs.

**Table 1.** An example of the initialization of 2 agents, each with 5 unique architectures and their costs.

| Max_layer_dims = [12288, 20, 5, 1] | | |
|---|---|---|
| No. | Leader_layers | Follower_layers |
| 1 | [12288, 7, 2, 1, 0.6931471805599453] | [12288, 4, 3, 1, 0.6931471805599453] |
| 2 | [12288, 7, 4, 1, 0.6931471805599453] | [12288, 6, 2, 1, 0.6931471805599453] |
| 3 | [12288, 15, 3, 1, 0.6931471805599453] | [12288, 1, 4, 1, 0.6931471805599453] |
| 4 | [12288, 8, 5, 1, 0.6931471805599453] | [12288, 2, 4, 1, 0.6931471805599453] |
| 5 | [12288, 19, 3, 1, 0.6931471805599453] | [12288, 11, 5, 1, 0.6931471805599453] |

## 2.5. Creating a new solution

New solutions (off_layers) are continuously generated in each iteration based on the list composed of architectures from the leader_layers and follower_layers. Function 1 initializes a new solution to build the new layer architectures. The first and last layers are taken directly from the first and penultimate columns of the unique layers, which are fixed. The middle layers are processed in a loop. For each middle layer, there is a probability Consideration Rate (CR) of selecting a value from the same column. A high CR, such as 0.9, indicates that the algorithm relies heavily on existing reasonable solutions, promoting exploitation. This helps refine known good configurations but might limit exploration. The selected indices are performed per column (layer) rather than for the entire architecture, promoting independence between layers, avoiding propagating changes across all layers simultaneously, and reducing the risk of disrupting promising architectures. This selection method closely resembles Harmony Search (HS) [19]. However, rather than using a stochastic random selection of individuals from the entire Harmony Memory (HM) population and the worst solutions are removed [20,21], Function 1 utilizes the probability of selection, known as Roulette Wheel Weighting [22,23], which is calculated from the cost of the solution rather than its rank in the population, to enhance the likelihood of choosing higher-fitness configurations. If selected, weighted probabilities based on cost are used to pick an index. The higher-fitness solutions (those with lower costs) will have a greater chance of being selected while maintaining diversity through stochastic selection. Then, with the probability Pitch Adjustment Rate (PAR), the value is adjusted by adding a random integer between -2 and 2. With PAR = 0.3, there is a 30% chance that a selected value undergoes a tweak. Finally, the adjustment is clipped to ensure it stays within bounds. If not selected from unique_layers, it generates a random integer within the allowed range. Roulette Wheel Weighing computes the selection probability for each solution using equations 3 and 4 [22].

$$P_i = \frac{fitness_i}{\Sigma_{j=1}^{N} fitness_j} \quad (3)$$

Where N is the total number of solutions, and for minimization problems (the lower cost = better):

$$fitness_i = \frac{1}{1 + cost_i} \quad (4)$$

**FUNCTION 1: GENERATION OF A NEW SOLUTION**

```
Input: unique_layers , CR, PAR, max_layer_dims
Output: new_layer_dims
Extract costs and calculate weighted probabilities for selection
# Exclude the first and last column
Middle_layers = composed_layers[:, 1:-1]
FOR i in range(middle_layers.shape[1]):
    IF random.random() < (CR = 0.9):
        Select an index from the current column using weighted probabilities
        IF random.random() < (PAR = 0.3):
            x += random.randint(-2, 2)
            Clip to bounds
    ELSE: # Random generation
        x = random.randint(1, max_layer_dims[i +1] +1)
    New_solution.append(int(x))
new_layer_dims = [int(first_layer_value[0])] + new_solution
Return new_layer_dims
End
```

## 2.6. Pareto optimization

In multi-objective optimization (MOO), a single solution that optimizes all objectives simultaneously rarely exists due to conflicting goals. Instead, MOO identifies a set of trade-off solutions known as Pareto-optimal solutions. These solutions represent the best possible compromises: if reallocating resources cannot improve one cost without increasing another, then the solution is optimal. Dominance ensures Pareto-optimal solutions are incomparable—no solution dominates another on the Pareto front. Solution A is said to dominate solution B if two conditions hold [24,25]:

- A is at least as good as B in all objectives. $f_1(A) \leq f_1(B)$
- A is strictly better than B in at least one objective. $f_1(A) < f_1(B)$

In the proposed method, Function 2 is introduced, an alternative approach that may not accurately implement the standard Pareto ranking (as in NSGA-II) [26]. Instead, it assigns a rank based on how many solutions in the entire population dominate the solution. The rank is the count of dominators. This approach uses Pareto ranking to filter out the worst solutions but still retains some diversity by not eliminating all dominated individuals immediately. If only non-dominated solutions are kept, the population might lose diversity early. By allowing some unique solutions to stay, the algorithm can explore different regions of the solution space. The proposed method's rank may not accurately capture the hierarchical front structure, resulting in different rank numbers. However, the overall order, when sorted by rank and cost, might still align with the standard front order. For example, solutions in front 1 (rank 0) come first, followed by solutions in front 2 (rank 1), and so on, maintaining the same sequence even if the rank numbers differ. Sorting by Pareto dominance rank followed by cost ensures that the solution with the lowest cost will always be ranked first in the list. Because a solution with the lowest cost is non-dominant, it will always have a Pareto dominance rank of 0. If sorted by cost, it will always be ranked first in the list. Moreover, Pareto dominance sorting helps retain solutions with smaller matrix dimensions. For example, a solution with layer_dims = [12288, 3, 4, 1] is prioritized over [12288, 4, 4, 1]. However, if solution 2 has a lower cost, it may dominate solution 1 despite its larger size. Therefore, this method achieves an optimal balance between low-cost and small-size configurations.

**FUNCTION 2: PARETO DOMINANCE RANK**

```
Input: agent_layers
Output: Pareto dominance rank
Popsize = number of solutions in agent_layers
Rank = a list to store the rank of each solution
Dominated = list to track how many solutions dominate each solution
FOR i in range(popsize):
    FOR j in range(i+1, popsize):
        IF solution[j] dominates solution[i]:
            Dominated[i] += 1
        ELIF solution[i] dominates solution[j]:
            Dominated[j] += 1
    Rank[i] = dominated[i]
Return Rank
End
```

In the proposed function, after the inner loop over j, the rank[i] is assigned to dominated[i]. Dominated[i] is the count of how many solutions dominate solution i. So, the rank here is set to the number of solutions that dominate it. In Pareto dominance ranking, a lower rank (like 0) means it's non-dominated. So, if a solution is not dominated by anyone, its rank would be 0 (Pareto front). Then solutions dominated by one solution would have rank 1, etc. The proposed function's approach uses the number of dominators as the rank, so the lower is better. For each i, we compare with j from i+1 to popsize. When i is 0, j starts at 1. Then, when i=1, j starts at 2. So, each pair is compared once. In a standard dominance comparison, if j dominates i, then i is dominated by j. Similarly, if i dominates j, then j is dominated by i. For each pair, both possibilities are checked because dominance is a pairwise comparison.

## 2.7. Pairing and Mutation

Function 3 performs row-based crossover between two parent neural network parameter sets to generate two child networks. It uses vectorized operations for efficiency, with distinct strategies for weights (W) and biases (b). In GA, a chromosome represents a candidate solution. Here, the neural network parameters (weights and biases) form the chromosome, with weights as matrices and biases as vectors. Each row of a weight matrix (W) is analogous to a gene segment, allowing fine-grained crossover operations at the row level. For each row in a weight matrix, a random crossover point is generated. Elements before the crossover point in a row come from leader_params, and elements after come from follower_params. In Figure 3, the off_params_1 combines the Leader's left and the Follower's right segments. Off_params_2 does the opposite. This mirrors the crossover in binary representations, where crossover shuffles existing bits (0 and 1) but cannot create new ones. In continuous spaces, this becomes more problematic because the search space is infinitely larger, and recombination alone cannot "fill gaps" between parent values. So, it depends on the mutation. Without mutation, the algorithm cannot escape the initial population's value. For example, if only crossover points are used, the initial population lacks a weight value of 0.3; crossover alone will never produce it. However, since the learning rate of neural networks is typically very small [27–29] (around 0.01 to 0.001, for example), the neural network parameters (such as weights and biases) are interdependent. Using linear blending might treat them as independent, leading to incoherent combinations that don't preserve the functional relationships needed. It could break the coordinated patterns necessary for feature extraction. Biases might be more tolerant to blending since they are additive parameters, but weights have multiplicative effects through connections. So, blending biases could still work, but blending weights might not. Biases are blended using a random weight (alpha) and a blended crossover (arithmetic crossover), where each element is a linear combination of parent values. In the blending method, two offspring are generated from the two parents by equations 5 and 6. For each element in the bias vector, a random alpha ∈ [0, 1] is randomly generated. The offspring's value is computed as:

$$b_{new1} = alpha * \ b_{leader} + (1 - alpha) * \ b_{follower} \tag{5}$$

$$b_{new2} = alpha * \ b_{follower} + (1 - alpha) * \ b_{leader} \tag{6}$$

**FUNCTION 3: CROSSOVER ROWS VECTORIZED**

```
Input: leader_params, follower_params
Output: off_params_1, off_params_2
# Iterate over all keys in the parameter dictionaries
FOR key in parameter.keys():
    IF key.startswith("W"):
        Generate random crossover points for all rows at once
        Generate off_params_1, off_params_2 by combining parts from Leader and Follower_ params
    ELIF key.startswith("b"):
        Generate a random alpha array with the same shape as b
        # Perform linear combination for biases for both Offspring
        off_params_1[key] = alpha * leader_params[key] + (1 - alpha) * follower_params[key]
        off_params_2[key] = alpha * follower_params[key] + (1 - alpha) * leader_params[key]
Return Output
End
```

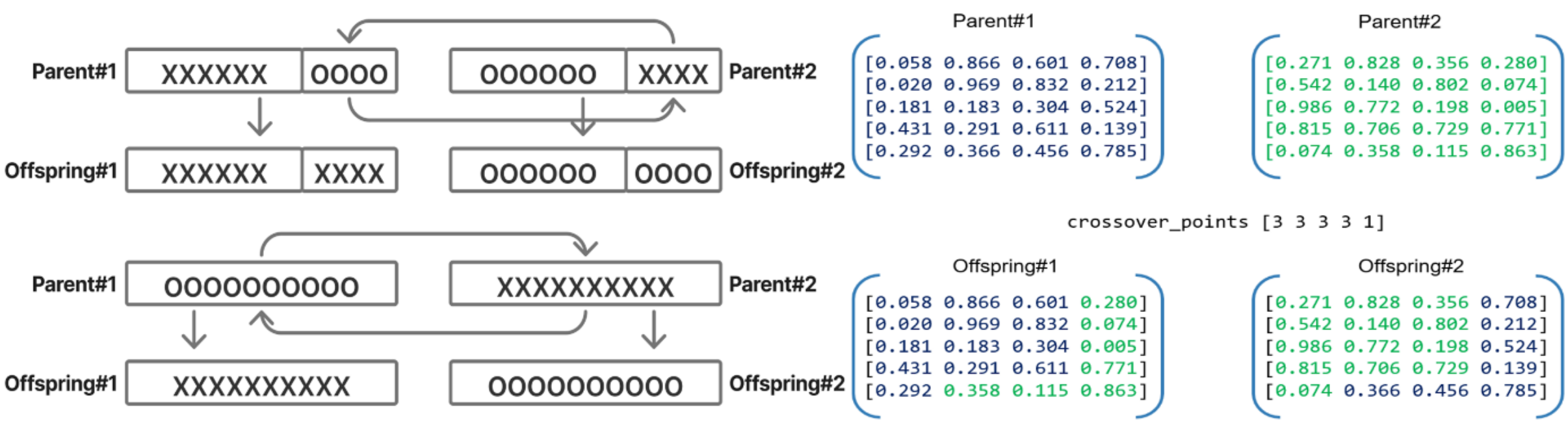

Figure 3. Pairing process of multi-point crossover

In Figure 4, the mutation parameters use a mutation rate decay (0.9 → 0.1 over 20,000 generations) and a mutation scale (0.008). The mutation rate starts at 0.9 (90% probability of mutating each parameter) and linearly decays to 0.1 (10%) over 20,000 generations. This is an adaptive mutation strategy that transitions from exploration to exploitation over time. The mutation scale (step size) determines the magnitude of perturbations applied to parameters. A fixed value of 0.008 means small, consistent parameter changes, adding Gaussian noise with $\sigma = 0.008$. For a weight value $W = 0.5$, a mutation might produce $W' = 0.5 + N(0, 0.008)$, where N is Gaussian noise. This ensures gradual, stable updates.

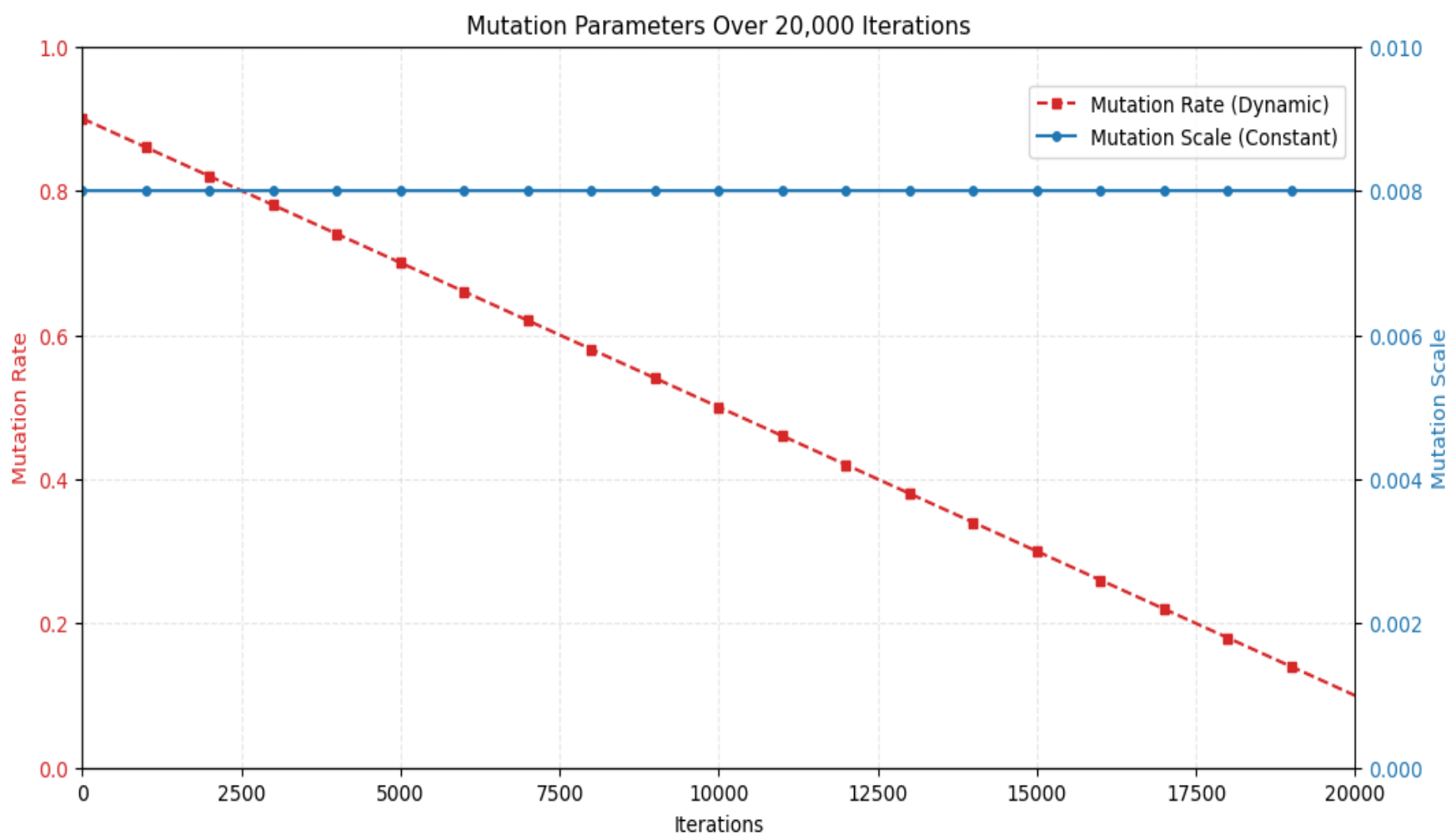

Figure 4. Mutation parameters over 20,000 iterations

Function 4 implements a Gaussian (normal distribution) mutation. It generates random perturbations from a Gaussian distribution with mean 0, scale mutation_scale for all parameters, but a mutation_rate determines whether a parameter is actually mutated. These perturbations (mutations) are added to the agent_params. Gaussian noise mimics stochastic gradient descent (SGD) noise, making it suitable for hybrid neuroevolution strategies.

**FUNCTION 4: MUTATION**

***Input:*** *agent_params, mutation_rate, mutation_scale*

***Output:*** *Mutated_params*

***FOR*** *key* ***in*** *agent_params.keys():*

```
        # Generate a mask for mutation
        Mask = np.random.rand(*parameters[key].shape) < mutation_rate
        # Generate random mutation values
        Mutations = np.random.normal(loc=0, scale=mutation_scale, size = agent_params[key].shape)
        Mutated = agent_params[key] + mask*mutations
        Mutated_params[key] = mutated
    Return Mutated_params
    End
```

## 2.8. Combining two offspring

Algorithm 2 merges two offspring parameters of the proposed method (off_params_1 and off_params_2), selects the best candidates using Pareto dominance and cost, and performs the following tasks:

- Merge and rank solutions: combines two sets of solutions (off_layers_1 and off_layers_2) into a single list, tagging each entry with its source ("1" for off_layers_1, "2" for off_layers_2) to track the origin of each solution to determine how to trim parameters later. These sets have the same solutions but different costs due to variations in the parameters. After joining, it assigns a rank to each solution based on Pareto dominance. Lower ranks indicate better solutions. Each entry in sorted_entries follows this structured format (pareto_dominance_rank, ([layer_dims, cost], source_tag)).
- Select unique solutions: ensures unique solutions architectures (based on layer dimensions) are retained.
- Update parameters: updates off_params based on the selected solutions for each unique solution, trimming unused portions and merging overlapping regions into off_params. Boolean masks are used to track updated regions in parameter matrices.
- Because solutions in off_layers are initialized as unique, merging off_layers_1 and off_layers_2 ensures the resulting updated_off_layers retains the same set of unique solutions (architectures) as initialized.

In Figure 5, the example demonstrates the merging of two offspring (off_layers_1 and off_layers_2) generated after 1000 iterations of pairing leader_params and follower_params. The experimental setup includes a neural network defined by the maximum possible values max_layer_dims = [50, 5, 5, 1], comprising an input layer (50 neurons), two hidden layers (with a maximum of 5 neurons each), and an output layer (1 neuron) for binary classification. Input X is a 50x100 matrix of random values representing 100 training examples with 50 features. Labels Y are binary labels (0/1) for 100 samples simulating a binary classification task. The focus is on the second hidden layer's weight matrix (W2), showing how solutions are selected, trimmed, and merged based on Pareto dominance and cost. Pairing leader_params and follower_params produces two offspring, off_params_1 and off_params_2. The second hidden layer's weight matrix (W2) for each offspring is extracted: W2_offspring_1 from off_params_1 and W2_offspring_2 from off_params_2. Solutions from off_layers_1 and off_layers_2 are combined into sorted_joined_layers, sorted by Pareto dominance and cost. Each entry in sorted_joined_layers follows the format: (pareto_dominance_rank, ([layer_dims, cost], source_tag)). The merging process is as follows:

- As the highest-ranked solution (Pareto rank 0), (0, ([50, 5, 3, 1, 0.19507], '2')) is prioritized. This leads to trimming 3 rows and 5 columns from W2_offspring_2 into W2_offspring.
- The second entry (0, ([50, 2, 5, 1, 0.31880], '2')) checks for uniqueness and adds 5 rows and 2 columns (green area), but skips overlapping regions that were already copied from previous entries.
- The third entry (0, ([50, 2, 2, 1, 0.50123], '1')) checks uniqueness and trims 2x2 from W2_offspring_1 but doesn't overwrite existing areas.
- The final W2_updated_offspring updates the cost based on this matrix. The merging loop might stop once 3 unique solutions (matching the size of off_layers_1) are selected.

**ALGORITHM 2: MERGE TWO OFFSPRING**

***Input:*** *off_params_1, off_params_2, off_layers_1, off_layers_2*

***Output:*** *off_params, off_layers*

*joined_layers = join off_layers_1 and off_layers_2 with source tags*
*Compute Pareto ranks from joined_layers*
*Sort joined_layers by Pareto rank and then by cost*
*Off_params = zero-initialized array with the same shape as off_params_1 or off_params_2*
***FOR*** *solution, source* ***in enumerate*** *(sorted_joined_layers):*
  *Layer_dims = tuple(solution[:-1]) # Extract layer dimensions exclude cost*
  ***IF*** *it is unique in off_layers:*
    *Off_layers.append(solution)*
    *Trim parameters from the corresponding source*
    *Copy trimmed parameters into off_params and create a mask*
    *(a Boolean array showing updated parts to exclude these parts for the next paste)*
*Ensure output sizes match the original agent_layers sizes*
***Return*** *Output*
***End***

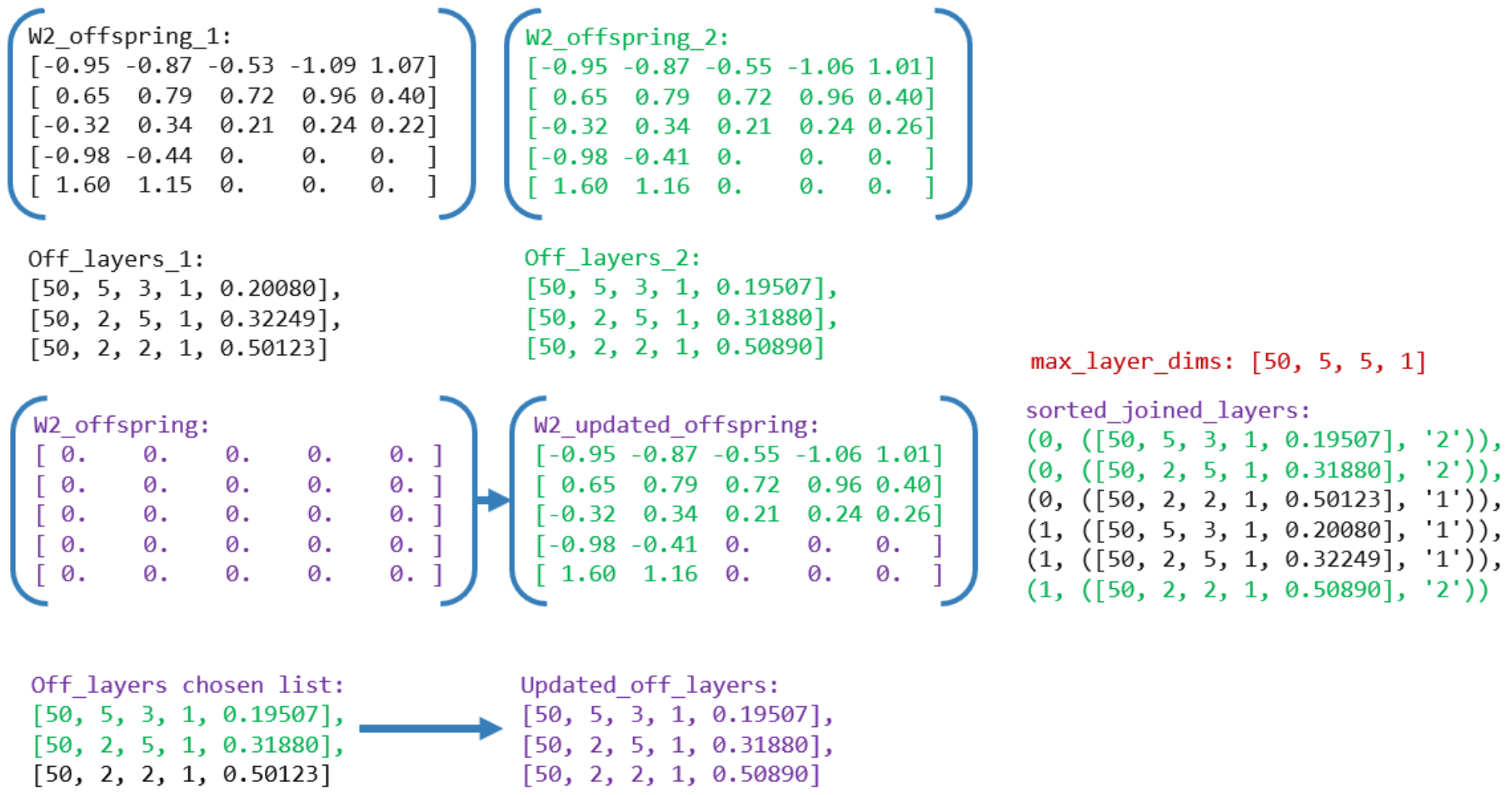

Figure 5. Dummy example of merging two offspring after 1,000 iterations.

## 2.9. Anabolism Process

Algorithm 3 outlines the function of merging and updating parameters for the proposed method. This function enables efficient co-evolution of two populations (Leader and Follower) with dynamic parameter adjustment and diversity preservation through agent layer sizes. Key tasks include:

- Merging solutions from the Leader, Follower, and new offspring parameters.
- Updating parameters (weights/biases) for Leader and Follower while preserving architectural constraints.
- Join solutions by joining leader_layers, follower_layers, and off_layers into joined_layers, tagging each entry with its source ("lead", "foll", or "off"). Compute Pareto dominance ranks (pareto_dominance_rank function) to prioritize non-dominated solutions. After that, track the seen layer configurations to avoid duplicates using seen_lead and seen_foll sets.
- Merge Parameters: For each solution, update leader_params and follower_params based on source (lead/foll/off). Alternating assignment (idx%2 splits solutions between Leader/Follower). Combines parameters from different solutions while avoiding conflicts using boolean masks (mask_trim). Merge parameters with mask_trim_lead and mask_trim_foll to avoid overwriting existing values and ensure overlapping regions (e.g., shared layer dimensions) are updated without corruption.
- Maintain agent layer sizes: If updated agent layers are smaller than the original sizes, generate new solutions (create_new_solution) to fill gaps. Trim updated agent layers to original sizes to ensure population stability.

The reason for the indices for leader_layers assigned as even index and follower_layers as odd index instead of splitting the indices into two halves (e.g., upper half for leader_layers and lower half for follower_layers) is that the proposed method's objective is not solely to find the best cost value but to identify the most suitable weight configurations. Assigning the best cost to the Leader and the second best to the Follower risks premature convergence, as the system overprioritizes exploitation at the expense of exploration. Furthermore, segregating layers into fixed groups (e.g., upper layers assigned to the Leader and lower layers to the Follower) could destabilize training, causing oscillations in layer dominance and hindering the discovery of optimal configurations. Thus, the even-odd layer allocation ensures a balance between exploitation and exploration: the Leader retains the best-performing weights for refinement, while the Follower explores alternative configurations to challenge and update the Leader dynamically.

**ALGORITHM 3: ANABOLISM PROCESS**

```
Input: leader_params, follower_params, off_params, leader_layers, follower_layers, off_layers
Output: updated_leader_params, updated_ follower_params,
        updated_leader_layers, updated_follower_layers
joined_layers = join leader_layers, follower_layers, off_layers with source tags
Compute Pareto ranks from joined_layers
Sort joined_layers by Pareto rank and then by cost
FOR idx, (solution, source) in enumerate (sorted_joined_layers):
  | Layer_dims = tuple(solution[:-1]) # Extract layer dimensions exclude cost
  | IF idx %2 ==0: # Even index → updated_leader_layers
  |   | IF it is unique in updated_leader_layers:
```

```
            updated_leader_layers.append(solution)
            Trim parameters from the corresponding source
            Copy trimmed parameters into leader_params  and create a mask
            (a Boolean array showing updated parts to exclude these parts for the next paste)
        ELSE:
            IF it is unique in updated_follower_layers:
                updated_follower_layers.append(solution)
                Trim parameters from the corresponding source
                Copy trimmed parameters into follower_layers and create a mask
    ELSE:# odd index → updated_follower_layers
        IF it is unique in updated_follower_layers:
            updated_follower_layers.append(solution)
            Trim parameters from the corresponding source
            Copy trimmed parameters into follower_layers and create a mask
        ELSE:
            IF it is unique in updated_leader_layers:
                updated_leader_layers.append(solution)
                Trim parameters from the corresponding source
                Copy trimmed parameters into leader_layers and create a mask
Ensure output sizes match the original agent_layers sizes
Truncate the updated_leader_layers and updated_follower_layers to the original length.
Return Output
End
```

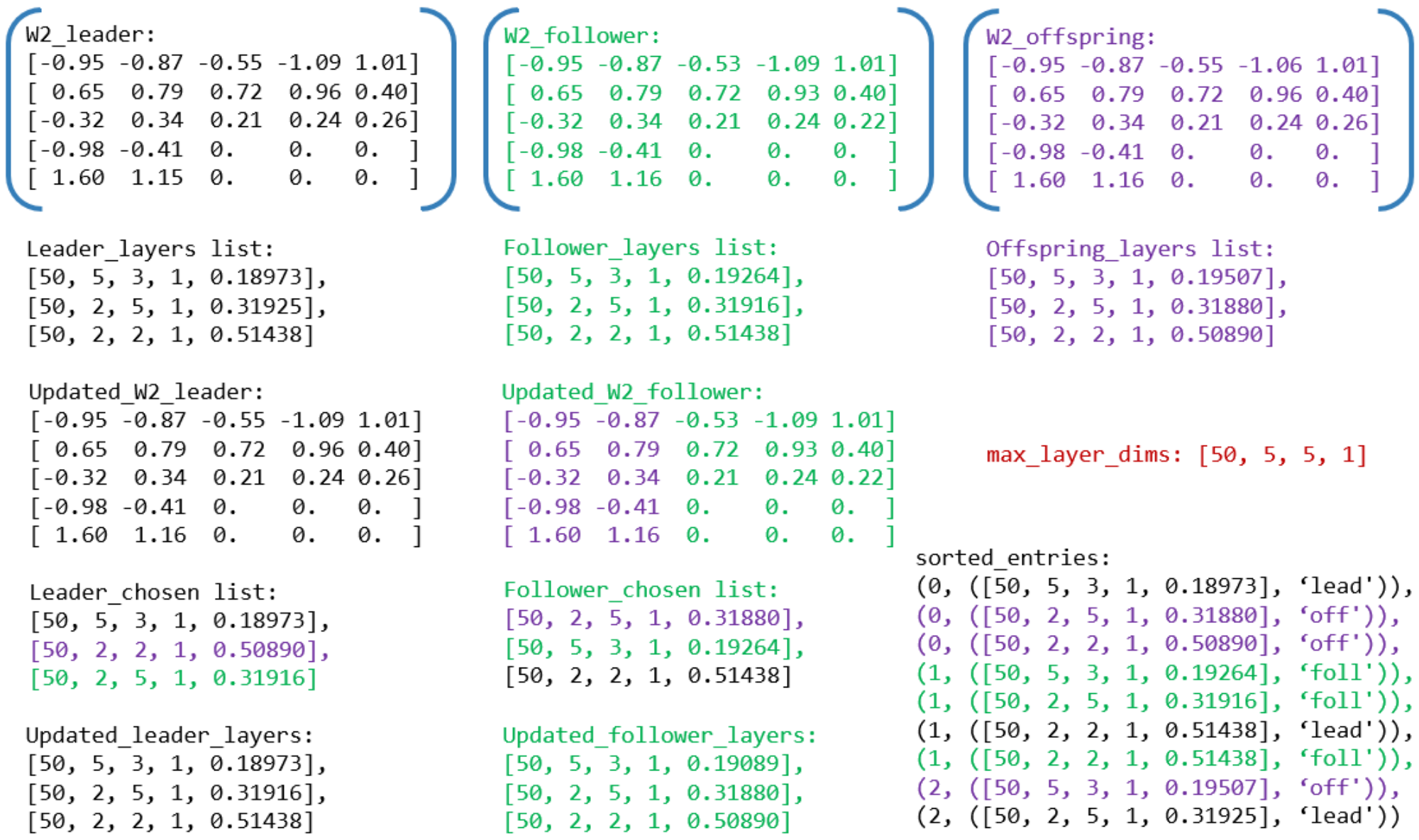

Figure 6. Dummy example of the Anabolism process after 1,000 iterations.

Figure 6 provides a schematic illustration of the Anabolism process after 1,000 iterations, using a dummy example for clarity. The merging process proceeds as follows, based on the sorted entries prioritized by Pareto rank and cost:

- Initial entry (idx=0): the first entry (idx=0) is assigned to leader_layers because idx%2==0. The entry has a "lead" tag, indicating it originates from the leader population. A 3x5 submatrix is trimmed from W2_leader (Leader's weight matrix) and pasted into the corresponding positions of updated_W2_leader (initialized as a copy of W2_leader). The updated regions in updated_W2_leader are marked in mask_trim_lead to prevent overwriting in subsequent steps.
- The second entry (idx=1) is assigned to follower_layers (idx=1 is odd). The source tag is "off" (offspring). A 5x2 submatrix is trimmed from W2_offspring (offspring weights) and pasted into the corresponding positions of updated_W2_follower (initialized as a copy of W2_follower), avoiding conflicts using a boolean mask (mask_trim_foll).
- The third entry (idx=2) is assigned to leader_layers (idx=2 is even). The layer dimensions ([50, 2, 2, 1]) are checked for duplicates in leader_layers. If not duplicated, a 2x2 submatrix from W2_offspring is pasted into updated_W2_leader. The solution is retained in the leader_chosen list. If duplicated, the merge is skipped for leader_layers and rechecked for follower_layers.
- The fourth entry (idx=3) is assigned to follower_layers. The source tag is "foll" (Follower). A 3x5 submatrix from W2_follower is pasted into updated_W2_follower in regions not previously masked.
- The fifth entry (idx=4) is assigned to leader_layers. The target positions in updated_W2_leader (from idx=0) are already masked. The merge is skipped, but the layer dimensions are not duplicated in leader_layers; the solution is copied to the leader_chosen list.
- The sixth entry (idx=5) is assigned to follower_layers. Layer Dims are checked for duplicates. The layer dimensions ([50, 2, 2, 1]) are not duplicated in follower_chosen list. Action is adding the solution for follower_layers.
- Seventh entry (idx=6) is assigned to leader_layers. Source Tag is "foll". The layer dimensions are duplicated in both leader_chosen and follower_chosen. Action is skipped entirely.
- The eighth entry (idx=7) is assigned to follower_layers. The source Tag is "off". The layer dimensions are duplicated in both leader_chosen and follower_chosen. Action is skipped entirely.
- Ninth entry (idx=8) is assigned to leader_layers. The source Tag is "lead". The layer dimensions are duplicated in both leader_chosen and follower_chosen. Action is skipped entirely.
- Cutting the updated_leader_layers and updated_follower_layers to the original length. Ensure output sizes match the original agent_layers sizes. If the size is smaller, create new solutions to match the original.

After processing all entries, the system updates the leader and follower populations. If follower_layers contains solutions with lower costs than the current Leader, roles are swapped: the best Follower becomes the new Leader, and vice versa.

## 3. RESULTS AND DISCUSSION

Table 2 compares the performance between the proposed method and gradient descent across different neural network architectures. The results are calculated using a provided prediction function. Once the model's cost falls below the stop_cost threshold, it generates two parameter sets, leader_params and follower_params, with 10 layer architecture configurations (5 for each set), ranked by Pareto dominance and cost. For networks with 2 and 3 layers (e.g., layer_dims = [12288, 7, 1] and [12288, 17, 4, 1]), the propsed method outperforms gradient, achieving train/test accuracies of 99.04% and 80%, respectively, compared to 99.52% and 74% for gradient descent with the same layer_dims = [12288, 17, 4, 1]. For a 2-layer neural network, layer_dims = [12288, 7, 1], the proposed method performs better, with train/test accuracies of 100% and 74%, compared to 100% and 72% for gradient descent. An interesting point is that the value of max_layer_dims ultimately becomes identical to the optimized layer dimension configurations after training. This confirms the proposed method's capability not only in optimizing parameters but also in automatically refining and selecting effective layer sizes. However, with a more complex 4-layer neural network, the proposed method underperforms compared to gradient descent despite achieving a higher training accuracy (99.52% vs. 98%); it reaches a smaller testing accuracy (76% vs. 80%). This indicates a classic overfitting problem. The model fits the training data extremely well, but generalizes poorly to unseen data [30,31]. This behavior is also observable in the layer dimension configurations and cost values of both the Leader and follower solutions. All 10 generated architectures (5 leaders and 5 followers) achieved extremely high performance on the training test, with nearly 100% accuracy, suggesting that the model has overlearned the training data patterns. This is because the proposed method experiences more overfitting than gradient descent. The proposed model learns too effectively, resulting in greater variance compared to gradient descent. The higher variance in the proposed method observed across all architectures is attributed to its search process, which involves random mutations and recombination, resulting in greater variability between runs compared to gradient-based optimization. While this can be beneficial for escaping local minima, it also increases the risk of overfitting if not properly constrained. Therefore, the increased tendency towards overfitting in the proposed method can be attributed to highly adaptive optimizations and a lack of regularization [32]. The evolutionary algorithm optimizes both weights and architectures aggressively, potentially leading the model toward overly complex configurations that capture noise in the training data. Unless explicitly controlled, the proposed method may focus solely on minimizing training costs, neglecting generalization.

The system used random.seed(42). Figure 7 displays the comparison of convergence between the proposed method and Gradient Descent. In Figure 7, the system only changes max_layer_dims, showing that the more layers there are, the harder it is for the system to converge, and it requires more iterations to reach the optimal cost value. Meanwhile, the Gradient descent converges faster, so it has a steeper slope. The optimal best values of the Leader and Follower will always follow each other throughout the optimization process.

Table 2. Comparison of proposed optimization and Gradient Descent on different Neural Network Architectures

| Max_layer_dims = [12288, 20, 5, 1], stop_cost = 0.035 | | | | |
|---|---|---|---|---|
| No. | Leader_layers | Train%/test% | Follower_layers | Train%/test% |
| 1 | [12288, 20, 3, 1, 0.03391] | 100/78 | [12288, 19, 3, 1, 0.03527] | 100/78 |
| 2 | [12288, 17, 4, 1, 0.05926] | 99.04/80 | [12288, 16, 5, 1, 0.06939] | 99.04/80 |
| 3 | [12288, 16, 4, 1, 0.06138] | 99.04/80 | [12288, 17, 3, 1, 0.08100] | 99.04/80 |
| 4 | [12288, 16, 3, 1, 0.07288] | 98.08/80 | [12288, 15, 3, 1, 0.09668] | 96.65/78 |
| 5 | [12288, 12, 2, 1, 0.32653] | 87.55/66 | [12288, 11, 2, 1, 0.38501] | 86.12/62 |
| Max_layer_dims = [12288, 7, 1], stop_cost = 0.015 | | | | |
| No. | Leader_layers | Train%/test% | Follower_layers | Train%/test% |
| 1 | [12288, 7, 1, 0.01495] | 100/68 | [12288, 7, 1, 0.01510] | 100/74 |
| 2 | [12288, 6, 1, 0.02269] | 99.52/64 | [12288, 6, 1, 0.02159] | 100/70 |
| 3 | [12288, 4, 1, 0.49732] | 85.16/52 | [12288, 4, 1, 0.50752] | 85.64/56 |
| 4 | [12288, 1, 1, 1.05136] | 65.07/40 | [12288, 1, 1, 1.13627] | 65.55/32 |
| 5 | [12288, 5, 1, 0.49743] | 85.16/52 | [12288, 5, 1, 0.50769] | 85.64/56 |

Max_layer_dims = [12288, 20, 7, 5, 1], stop_cost = 0.04

| No. | Leader_layers | Train%/test% | Follower_layers | Train%/test% |
|---|---|---|---|---|
| 1 | [12288, 13, 7, 5, 1, 0.03944] | 100/74 | [12288, 12, 7, 5, 1, 0.04106] | 100/74 |
| 2 | [12288, 13, 7, 4, 1, 0.04006] | 100/74 | [12288, 13, 6, 4, 1, 0.04210] | 100/74 |
| 3 | [12288, 12, 6, 4, 1, 0.04189] | 100/74 | [12288, 13, 5, 4, 1, 0.04487] | 99.52/74 |
| 4 | [12288, 12, 5, 4, 1, 0.04316] | 99.52/74 | [12288, 12, 7, 3, 1, 0.05081] | 99.52/74 |
| 5 | [12288, 13, 6, 3, 1, 0.05398] | 99.52/76 | [12288, 12, 6, 3, 1, 0.05276] | 99.04/74 |

Gradient Descent

| No | Layer_dims | Train%/test% |
|---|---|---|
| 1 | [12288, 7, 1, 0.048] | 100/72 |
| 2 | [12288, 17, 4, 1, 0.0378] | 99.52/74 |
| 3 | [12288, 20, 7, 5, 1, 0.092] | 98/80 |

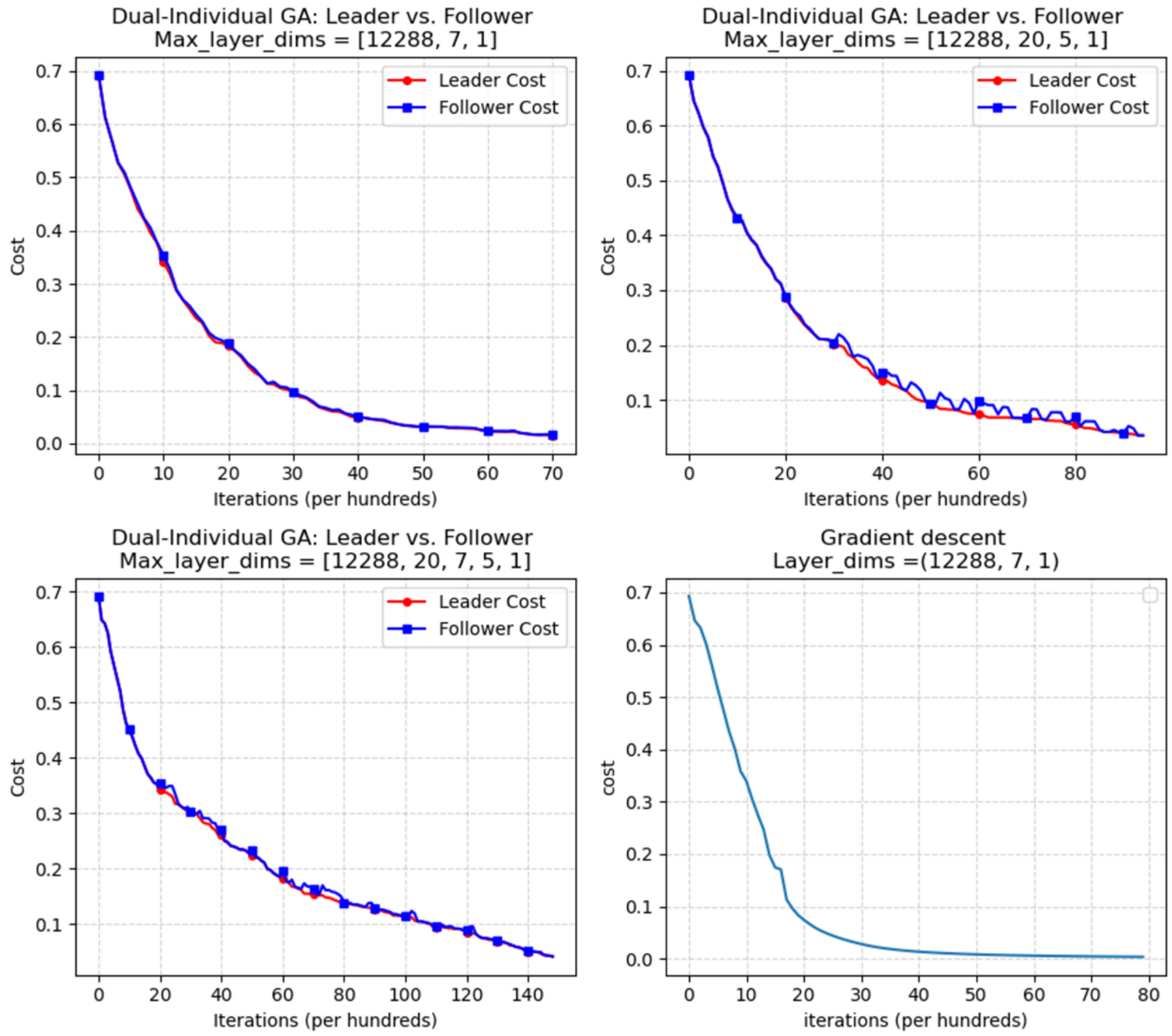

Figure 7. Training cost curves for Proposed Method vs. Gradient Descent

## 4. CONCLUSIONS

The proposed evolutionary algorithm (EA), Dual-Individual GA, introduces a promising new approach to optimizing architectures with a large number of parameters, such as neural networks, particularly in scenarios where gradient-based methods fall short. Despite demonstrating high performance, the proposed method still faces limitations, including a high number of iterations and susceptibility to overfitting. To address these challenges, future work plans to integrate the evolutionary algorithm with gradient-based methods, using the proposed method as the primary optimizer while leveraging gradient-based methods for local exploitation. This hybrid approach aims to reduce the number of iterations. Additionally, further enhancement will incorporate techniques such as weight averaging and L1/L2 regularization to improve generalization, stabilize training, and mitigate overfitting.

## FUNDING

This research was financially supported by Soongsil University, Seoul 156-743, Republic of Korea.

## INSTITUTIONAL REVIEW BOARD STATEMENT

The study did not require ethical approval.

## INFORMED CONSENT STATEMENT

Informed consent was obtained from all subjects involved in the study.

## DATA AVAILABILITY STATEMENT

The source code for this study is available upon request. The source code is provided as a Jupyter Notebook (.ipynb) file and can be executed in multiple environments, including Jupyter Notebook/Lab, and is compatible with Visual Studio Code (version 1.100.3, 2025). The code requires Python 3.11.9 for optima compatibility. To execute it, simply open the .ipynb file.

## ACKNOWLEDGMENT

This research was partly funded by the Technology Innovation Program (or Industrial Strategic Technology Development Program-Materials/Parts Package Type) (20016038, development of textile-IT converged digital sensor modules for smart wear to monitor bio & activity signal in exercise, and KS standard) funded By the Ministry of Trade, Industry & Energy (MOTIE, Korea) and Korea Institute for Advancement of Technology (KIAT) grant funded by the Korea Government (MOTIE) (P0012770).